\documentclass{article}

\usepackage{arxiv}

\usepackage[utf8]{inputenc} 
\usepackage[T1]{fontenc}    
\usepackage{hyperref}       
\usepackage{url}            
\usepackage{booktabs}       
\usepackage{amsfonts}       
\usepackage{nicefrac}       
\usepackage{microtype}      
\usepackage{lipsum}
\usepackage{graphicx}
\usepackage{booktabs}
\graphicspath{ {./images/} }
\usepackage{amssymb}
\usepackage{bbding}
\usepackage{pifont}
\usepackage{wasysym}
\usepackage{utfsym}
\usepackage{fontawesome}
\usepackage{enumitem}
\usepackage{multirow}
\usepackage{multicol}

\title{Knowledge Prompt Chaining for Semantic Modeling}

\author{
 Ning Pei Ding \\
  College of Informatics\\
  \texttt{Huazhong Agricultural University} \\
   \And
 Jing Ge Du \\
  College of Informatics\\
  \texttt{Huazhong Agricultural University} \\
  \And
 Zai Wen Feng \\
  College of Informatics\\
  \texttt{Huazhong Agricultural University} \\
}

\begin{document}
\maketitle
\begin{abstract}
The task of building semantics for structured data such as CSV, JSON, and XML files is highly relevant in the knowledge representation field. Even though we have a vast of structured data on the internet, mapping them to domain ontologies to build semantics for them is still very challenging as it requires the construction model to understand and learn graph-structured knowledge. Otherwise, the task will require human beings' effort and cost. In this paper, we proposed a novel automatic semantic modeling framework: Knowledge Prompt Chaining. It can serialize the graph-structured knowledge and inject it into the LLMs properly in a Prompt Chaining architecture. Through this knowledge injection and prompting chaining, the model in our framework can learn the structure information and latent space of the graph and generate the semantic labels and semantic graphs following the chains' insturction naturally. Based on experimental results, our method achieves better performance than existing leading techniques, despite using reduced structured input data. Our code is available at \url{https://github.com/dingningpei/LLM_Semantics}.
\end{abstract}


\section{Introduction}
Currently, structured data sources remain the primary choice for storing enterprise or web data. However, a vast amount of structured data on the internet exists in different formats, making the integration of heterogeneous data sources a critical topic for research to promote data sharing \cite{dhamankar2004imap}. Domain ontologies offer a solution to the challenge of integrating heterogeneous data sources. As a hierarchical knowledge representation model, a domain ontology defines classes and properties specific to a domain to describe factual knowledge about the real world. Mapping structured data sources to a domain ontology produces a semantically meaningful structured model, referred to as the semantic model of structured data sources, and the process is known as semantic modeling. Semantic modeling enables the integration of heterogeneous data from various sources and includes two main steps. The first step is semantic labeling, which involves annotating attribute columns in the data source using classes and data properties from the domain ontology to derive semantic types for the target source. Based on this annotation, the second step is to identify and establish semantic relationships between attributes in the data source using object properties in the domain ontology. This step typically requires incorporating the actual content of the structured data to develop a reliable semantic model. Our research focuses on a challenging aspect of this task. For example, in a domain ontology, there might exist a potential triple: \textit{<Event, had\_participant, Actor>}. Since the class \textit{Activity} is a subclass of \textit{Event}, the triple \textit{<Event, had\_participant, Actor>} can be further refined into the triple \textit{<Activity, had\_participant, Actor>}. The refined triple remains valid. During the semantic modeling process, whether to refine a potential triple and, if so, into which triple, depends on specific reasoning in different situations. This ontology reasoning process is crucial for the semantic modeling task as it directly impacts the reliability of the semantic model.

In previous studies on automatic semantic modeling for structured data sources \cite{Taheriyan2016LearningTS}\cite{de2018machine}\cite{10.1145/3308558.3313711}\cite{futia2020semi}, researchers applied various methods to this task, ranging from Steiner trees to probabilistic graphical models and graph neural networks. These methods often required constructing integrated graphs through ontologies and known semantic models or utilizing knowledge graphs to train models for automatic semantic modeling, which demanded significant human effort or time. With the advent of LLMs, automatic semantic modeling tasks have embraced a more optimal solution.


In this paper, we introduce Knowledge Prompting Chaining, a novel framework for semantic modeling. First, our framework sterilizes the domain ontology, structure data given, and its corresponding semantics. All this knowledge will transform into the system prompt for Long In-context Learning(\textbf{LongICL}) instead of building integrated graphs as prior works did.
In the process of constructing the system prompt, we only adapt few points for each data given. For example, in tabular data, we will only apply three lines of data instead of the whole table.
The semantics are converted into a two-step solution: semantic labeling and semantic modeling. It helps LLMs solve the task following this two-step solution, which can naturally benefit our prompting chaining architecture. When the user inputs a new source, our framework will enter the chain1 part, in which we will serialize the new source and inject it into a semantic labeling prompt to generate semantic labels for the new source. This semantic label will be the input of the Chian2 part of our framework to finally output the semantic models of the new source by continually prompting the models in our framework. Our key contribution are: First, by inject graph-based knowledge into prompts properly, our work enhance LLMs' ability of in-context learning about structure information and latent representation of nodes in the graphs(domain ontolgoies and semantics). Secondly, we evaluated our methods on three real-world datasets. Our method outperforms state-of-the-art frameworks and underscores the benefits of implementing prompt chaining and pruning. Lastly, our method is high efficient for token usage. We only have to use a small part of the data points of structure data to generate its semantics, while prior works have to deal with the whole tabular, json and XML files.

\section{Related Work}
\paragraph{Automatic semantic modeling of structured data.}
Many studies focus on automatic semantic modeling of structured data. For source attribute semantic labeling, the first step in semantic modeling, DSL \cite{pham2016semantic} is a domain-independent automatic semantic labeling method based on machine learning, DINT \cite{rummele2018evaluating} combines feature engineering and machine learning to realize semantic annotation, and Meimei \cite{takeoka2019meimei} is an efficient probabilistic method to conduct semantic annotation in probability calculation through a multi-label classifier. Additionally, other work focuses on parsing the semantic relationships between source attributes. Early research \cite{Taheriyan2016LearningTS}\cite{de2018machine} used Steiner tree-based methods for automatic semantic modeling, but these were limited by the inability to correctly identify multiple edges with identical weights in the semantic model search graph. PGM-SM \cite{10.1145/3308558.3313711} employs probabilistic graphical models to score and rank the most reasonable semantic models. Deep learning methods have also been applied to this task. SeMi \cite{futia2020semi} uses graph neural networks to learn and mine semantic features from background-related data and predict potential semantic relationships in the target data sources. Our research uses advanced large language models and prompt engineering techniques for the automatic semantic modeling of structured data sources. The combination of LLMs and prompts provides powerful reasoning capabilities. At the same time, users can customize their requirements by simply modifying instructions, enabling flexible and interactive semantic modeling capabilities.

\paragraph{Large language models and prompts for structured data.}
LLMs have demonstrated outstanding performance in handling various text tasks, which has led researchers to apply them in the field of structured data. Since LLMs are sequence-to-sequence models, structured data must be converted into text format before it can be fed into LLMs \cite{jaitly2023towards}. \cite{singha2023tabular} found that DFLoader and JSON formats outperform other formats in all table transformation tasks. Researches like TabLLM \cite{hegselmann2023tabllm}, TAP4LLM \cite{sui2023tap4llm}, and GReaT \cite{borisov2022language} have applied LLMs to tasks such as table classification, table data understanding, and table data generation. Although LLMs have been successfully applied in structured data tasks, complex reasoning remains a challenge. Chain-of-Thought (COT) \cite{wei2022chain} is a prompt technique in the era of large language models that achieves good reasoning performance on complex tasks through a series of intermediate reasoning steps. \cite{zhao2023large} addressed question-answering tasks in complex tables by reformatting the table data structure and designing appropriate COT prompts. CHAIN-OF-TABLE \cite{wang2024chain} framework explicitly incorporates table data into the reasoning chain as an intermediate reasoning agent to guide LLMs in performing step-by-step table operations and updates. Knowledge Prompt is a method of embedding knowledge into prompts, and there have been studies applying Knowledge Prompt to tasks such as knowledge graph completion \cite{wei2024kicgpt}, multi-hop link prediction in knowledge graphs \cite{shu2024knowledge}, and knowledge-based VQA \cite{wang2024soft}. Our research focuses on using LLMs and prompts to achieve automated semantic modeling of structured data.

\section{Problem Statement}
\subsection{Overview}
Suppose we have a set of Ontologies ${\Omega}$ and a set of structure data $D = \lbrace (X_{i}^{c}, S_{i}) \rbrace_{i = 1}^{n}$,  where semantic $S_i \in \Omega$, $X_{i}^{c} = \lbrace (x_{a_j}^{r})_{j = 1}^{J_i}\rbrace_{r = 1}^{R_i}$, and $ c \in C$ for a set of structure format as XML, Json, or CSV. $n$ is the size of the data $D$ and $x_{a_j}^{r}$ is the value of the attribute $a_j$ in the row $r$. $J_{i}$ is the size of the attributes in the $ith$ data source, and $R_i$ is the size of the rows in the $ith$ data source. Semantic Modeling is to map the structure data $X$ to the target graph $S$ in the Ontologies $\Omega$. Semantic Modeling should include two parts: 1. Semantic Labeling, 2. Semantic Graph Building.

\subsection{Semantic Modeling}
\paragraph{Semantic Labeling} Semantic Labeling is to map function that map the attributes $a$ in the data source $X$ to the nodes in the Ontology $\Omega$. We use the triple set relation $N_{i} = \lbrace(a_j, p_j, n_j) \rbrace_{j = 1}^{J_i}$ as the mapping process of one certain data source, where $p_j$ is the properties of the target Ontology and $n_j$ is the node of the target Ontology and the pair $p_j, n_j$ is called the \textbf{ semantic type} of the attribute $a_j$.
The map function denoted it as $F: N_{i} = f(X_i, \Omega)$.

\paragraph{Semantic Graph Building} Semantic Graph Building is to find the construction path in the target Ontology, which can connect all the \textbf{Semantic Type}. We use the trips set $S_i = \lbrace{(e_g, p_g, e_g)}\rbrace_{g = 1}^{G_i}$ to describe the final semantic graph, where $G$ is the size of the triple set $S_i$. We use function $Q$ to demonstrate the process of finding the construction path. $Q: S_i = q(X_i, N_i, \Omega)$

Our goal is to find the approximate  $Q^*$ and $F^*$ function based on $\Omega$ and $D$. Given a new dataset $X^*$, the precision and recall scores between the predicted semantics $S^*$ and the golden semantics of $X^*$ should be as high as possible, where $S^* =  Q^*(X^*, F^*(X^*, \Omega), \Omega)$.


\section{Knowledge Prompt Chaining}
\subsection{Overview}

\begin{figure}[t]
\includegraphics[scale=0.1]{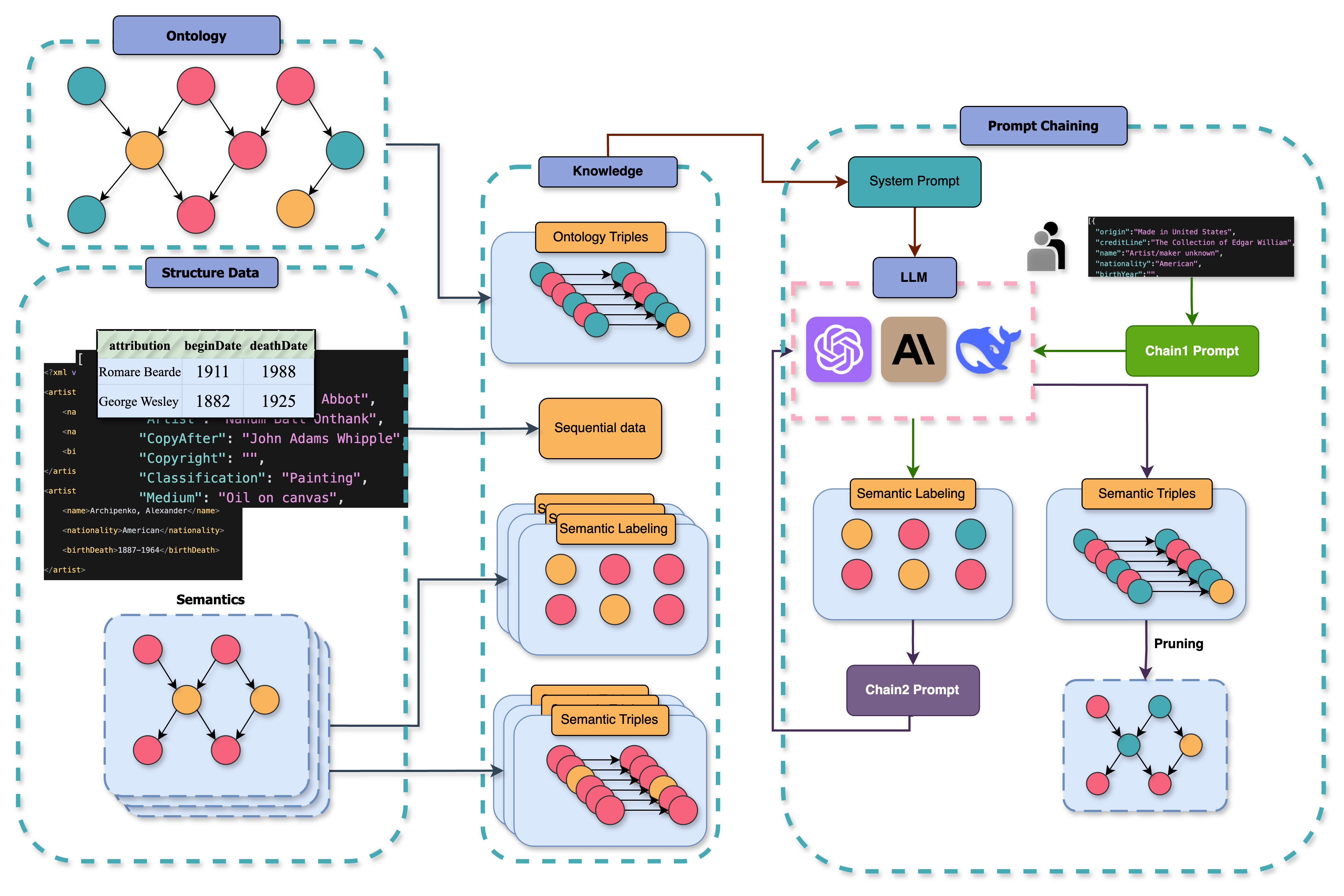}
\centering
\caption{An overview of Knowledge Prompt Chaining Framework}
\label{overview}
\end{figure}

Our Knowledge Prompt Chaining Framework is illustrated in Figure~\ref{overview}. The framework consists of two main parts:

1. Knowledge Integration: Our method adopts the Long-Context Large Language Model as Construction Function $F$ and $Q$. The data set and ontology are taken as knowledge input of the LLMs with hashtag $<Examples>$ and $<Ontology>$ correspondingly. All this knowledge is sterilized into sequential data and integrated into the knowledge prompt template as the system prompt of our automatic semantic modeling system.

2. Prompt Chaining: When the user inputs new structured data, we inject it into the designed prompt chaining process and output the semantic labeling with the hashtag $<Step1>$ first. The semantic labeling and data itself are taken as the input of the next step of the chain. It finally produces the semantic graph with hashtag $<Step2>$.



\subsection{Knowledge Integration}
Except for taking advantage of prior knowledge in the LLMs, we want to inject external knowledge into the prompt. However, the external knowledge base, $D$ and $\Omega$ in this task is the graph structure. So, the big challenge here is to inject the graph-based knowledge into a text-based prompt.

\paragraph{Structure Data Serialization} For the aim of using LLM, all the data input into LLM should be transformed into nature language representation. We denote this process as function $Serialize$ that transforms the structure data into text format data.
In the previous work\cite{10.1145/3308558.3313711}\cite{Taheriyan2016LearningTS}, they applied an encoder-focused serialization method to process the structured data. Typically, they convert the structured data into plain text, which a machine learning or deep learning model is trained on as a decoder. Our method focuses on text-based serialization\cite{fang2024large}. According to Singha's work \cite{singha2023tabular}, JSON format and Pandas Dataloader outperform other text-based serialization methods in prediction tasks. However, pandas dataloader cannot process structured data like XML. So in this task, we applied JSON format to sterilize $X$ and $S$. We only select three data points in each data as the high cost of token usage in LLMs.

\paragraph{Graph Serialization} Besides the structure data, the input of the LLM also include the graph data $\Omega$. We denote this Serialization as $GraphSeialize(\Omega)$. In this work, we process ontologies in JSON format to unify the input formats in LLM.

\begin{figure}[t]
\includegraphics[scale=0.3]{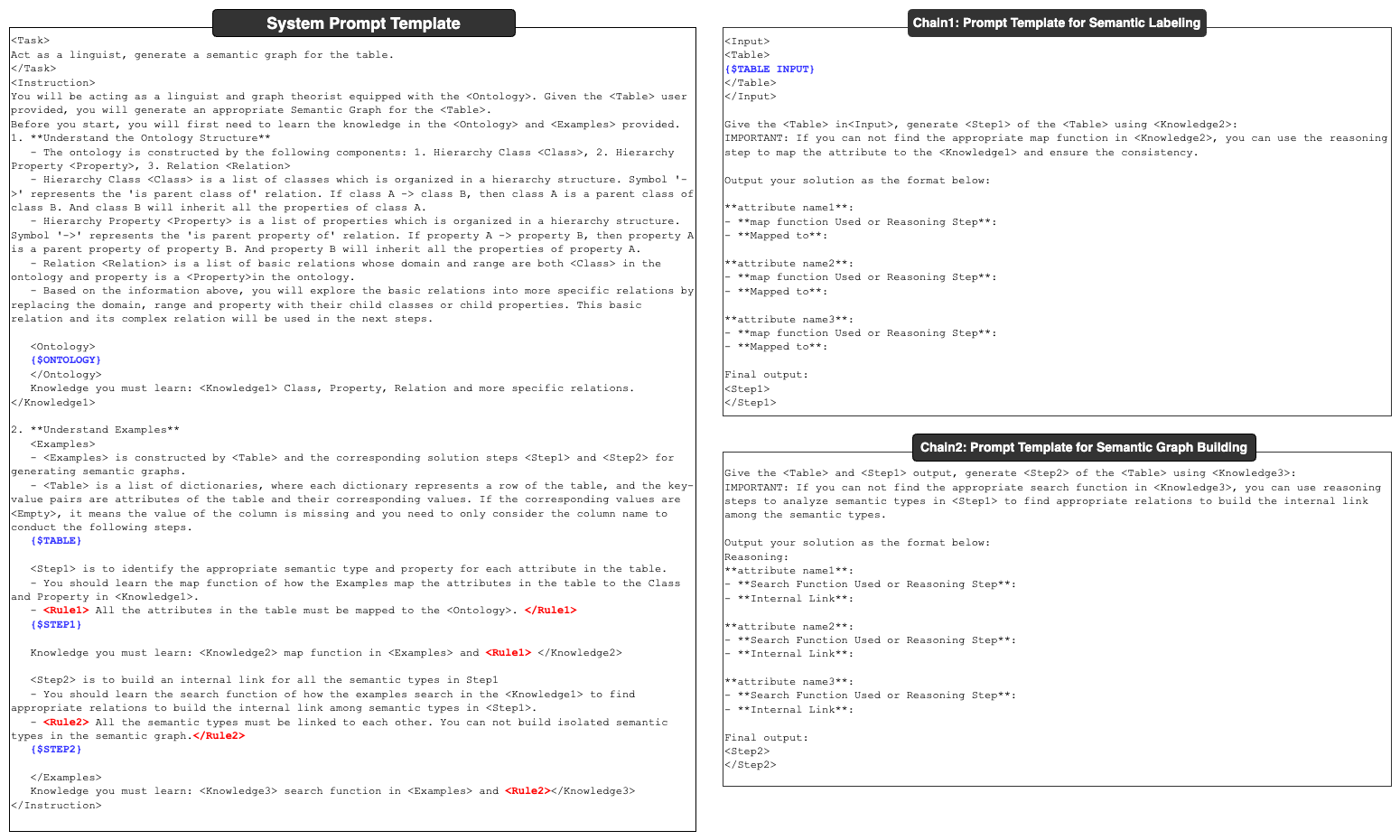}
\centering
\caption{Prompt Templates for prompting LLMs}
\label{template}
\end{figure}

When finishing the serialization, we get the processed data: $Serialize(X)$, $Serialize(S)$ and $GraphSeialize(\Omega)$. We then inject this data into our designed system prompt template in Figure~\ref{template}. $\{\$TABLE\}$ should be replaced with $Serialize(X)$ and $\{\$ONTOLOGY\}$ should be replaced with $GraphSeialize(\Omega)$. We divide  $Serialize(S)$ into two part: Semantic labels $Step1$ and semantic models $Step2$, which are used to replace $\lbrace{\$STEP1}\rbrace$ and $\{\$STEP2\}$ correspondingly. Now we can construct our final dataset for context learning: $$\\(\Omega, \\(Serialize(X_{i}), Step1_{i}, Step2_{i}\\)\\)$$

LLMs would later learn these two-step solutions to our task through context learning. In each $<Step>$, we add $<Rule>$ to reduce hallucinations in LLMs. The $<Rule>$ can be generated by machine-learning-based methods or written by experts manually. In this paper, $<Rule>$s are written by experts. Then the final system prompt is completed and fed into LLMs.

\subsection{Prompt Chaining with Pruning}
When a user inputs a new dataset $X^{*}$, we use the function above $Serialize$ to process it. We then prompt the LLM to generate semantic labels $N^{*}$ for $X^{*}$, which is chain1 part of the system. We do this prompt by constructing chain1 data as $(<Chain1 Prompt>, Serialize(X_{i}^{*}))$. Once we get semantic labels of the dataset, we enter the chain2 part: semantic graph building. We input chain2 data to LLM continually and build semantic models for $X^{*}$. The chain2 data is composed of $N^{*}$ from last step and $<Chain2 Prompt>$ in Figure~\ref{template}. We do not have to input $X^{*}$ here because we already fed it into LLM in chain1 part. In both $<Chain1 Prompt>$ and $<Chain2 Prompt>$, we do require LLM to give the detailed reasoning process of the output instead of outputting the result directly. This is inspired by Chain-of-Thought technique\cite{wei2022chain}. The whole Prompt Chaining process is denoted as:
$$
N^{*} =  LLM(<Chain1 Prompt>, Serialize(X_{i}^{*}))$$
$$S^{*}_{raw} = LLM(<Chain2 Prompt>, N^{*}))
$$

The final output of prompt chaining is graph-structured semantic models. It means we can use the pruning method on the graph to remove the wrong nodes in the output that are created by hallucinations of the LLMs. We check each node in the output and prune the node which cannot connect to attributes $a^{*}$ in $X^{*}$.

\section{Experimental Setup}
\subsection{Datasets}
We evaluated our method on three datasets, each of which consists of a set of structured data sources and an ontology that will be used to model the data sources. The first dataset, $ds_{crm}$, models 28 museum data sources using the CIDOC Conceptual Reference Model (CIDOC-CRM), which contains data from different art museums in the US. The second dataset, $ds_{edm}$, models museum data sources using the Europeana Data Model (EDM). The third dataset, $ds_{schema}$, models 12 football data sources using an extension of Schema.org. Table 2 provides more details on the evaluation datasets. Since this study uses LLMs to model structured data sources, the necessary preprocessing and cleaning were performed on the datasets to ensure that the LLMs could correctly recognize and process the data. The data processing details are as follows:

\begin{table}[!htbp]
\centering
\caption{The evaluation datasets $ds_{crm}$, $ds_{edm}$, and $ds_{schema}$.}
\label{tab:tableTab1}
\renewcommand{\arraystretch}{1.15}
\begin{tabular}{lccc}
\toprule
& $ds_{crm}$ & $ds_{edm}$ & $ds_{schema}$ \\
\midrule 
\#data sources & 28 & 28 & 12 \\
\#classes in domain ontologies & 84 & 52 & 9 \\
\#properties in domain ontologies & 288 & 248 & 19 \\
\#nodes in the gold-standard models & 634 & 445 & 135 \\
\#data nodes in the gold-standard models & 276 & 311 & 97 \\
\#class nodes in the gold-standard models & 358 & 134 & 38 \\
\#links in the gold-standard models & 595 & 417 & 123 \\
\bottomrule
\end{tabular}
\end{table}

\begin{itemize}
\item Table Data: The raw tabular data existed in various formats (CSV, XML, and JSON), which were converted into a unified JSON format. The content of the tables was stored in a unified list ([]), with each record in the original table corresponding to a dictionary object ({}) in the list. Due to the token input limitations of LLMs, only three records from each original table were retained in the processed data. Each dictionary represents a single record in the table, with keys as attributes (column names) and values as the corresponding data. If a value is missing, it is represented as $<Empty>$, but this does not affect the processing logic of the data.
\item Ontology Data: The ontology data were also processed into a JSON structure, organized into three main sections in the form of dictionaries. The $Nodes$ field lists the classes in the ontology, and the $Properties$ field lists the properties of the ontology, where each class and property may include its parent class or property, with inheritance relationships denoted by the "->" symbol. The $Potential\ triples$ field lists potential triples representing relationships between classes and properties.
\item Semantic Model Data: The semantic model data were also processed into JSON format, consisting of $semantic\_triples$ and $internal\_link\_triples$. Specifically, the $semantic\_triples$ field contains semantic triples that represent the relationships between table attributes and ontology nodes. The $internal\_link\_triples$ field contains internal link triples that represent the linking relationships between different entities within the ontology.
\end{itemize}
Please note that we have updated the semantic models in the ground-truth datasets to ensure that they are more reasonable and interpretable. Detailed information can be found in Appendix A.

\subsection{Models}
In this study, LLMs are required to understand complex tabular data and perform semantic modeling under ontology constraints. Therefore, the selection of LLMs must take into account several factors, including their ability to process long contexts, token input limitations, performance characteristics, and cost-effectiveness. Based on these considerations, we selected three distinct LLMs for our experiments: Claude 3.5 Sonnet, GPT-4 Turbo, and DeepSeek-V2.5.

\subsection{Experimental Design}
We conducted experiments using the three LLMs on three datasets. The purpose of the experiments was to assess the effectiveness of different LLMs in semantic modeling tasks on structured data across various domains. We compared our approach with three semantic modeling systems: Taheriyan et al. \cite{Taheriyan2016LearningTS}, De Uña et al. \cite{de2018machine} (Serene), and Vu et al. \cite{10.1145/3308558.3313711} (PGM-SM). The specific experimental settings are as follows.

In the LLMs experiment section, during data loading and splitting, we first indexed the data files and controlled the random shuffling of the indices using the $random\_state$ parameter. We selected 2023 and 2024 as the random seeds to mitigate potential biases arising from the initial order of the data files, ensuring consistency in repeated experiments and assessing the stability of model performance. The $test\_size$ parameter was fixed at 0.5 to ensure that half of the data was always used as the test set. Additionally, the $size$ parameter was fixed at 3, meaning that only the first three records from each table data file were loaded. To evaluate the models' learning capabilities under varying amounts of input data, we implemented three different settings for the $num\_files$ parameter: the first setting loaded only one file, the second loaded one-quarter of the files, and the third loaded one-half of the files. These settings were referred to as \textbf{\textit{one-shot}}, \textbf{\textit{quarter-shot}}, and \textbf{\textit{half-shot}}, respectively. These files represent known semantic models. To assess the impact of prompt engineering, we also designed a prompt ablation study to measure how different prompts affected model performance. These experimental settings were repeated across all models and datasets.

In the comparative experiment section for different semantic modeling methods, we applied the \textbf{\textit{half-shot}} setting to the three methods. The dataset partitioning was consistent with that used in the LLMs experiment to ensure that the comparison of different methods was performed under the same experimental setup. Additionally, for the PGM-SM method, we applied the Serene semantic labeler.

\subsection{Evaluation Metrics}
The tasks in this study consist of two steps: semantic labeling and semantic graph building. The results of both steps are evaluated using unified $Precision$ and $Recall$ metrics. These metrics are calculated by comparing the predicted semantic annotations and semantic models with the gold standard models. Assuming the correct semantic model of the data source $X$ is $S$, and the semantic model learned by our approach is $S^*$, we define $Precision$ and $Recall$ as follows:
\[Precision=\frac{|S\cap S^*|}{|S^*|}\]
\[Recall=\frac{|S\cap S^*|}{|S|}\]\textit{}where $S$ represents the gold standard semantic model for the data source $X$, which contains the set of correct semantic triples; $S^*$ represents the semantic model predicted by our approach, which contains the set of learned semantic triples. When evaluating the results of semantic annotation, only the semantic triples that annotate the source attributes are considered. For evaluating the results of semantic graph construction, all semantic triples in the semantic model are taken into account.

\section{Results}
In this section, we evaluate the effectiveness of the Knowledge Prompt Chaining approach for semantic modeling through a series of experiments aimed at addressing the following key research questions:
\begin{itemize}[leftmargin=*, labelwidth=0pt, labelsep=0.5em]
    \item[\textbf{•}] \textbf{Q1:} How does the Knowledge Prompt Chaining approach perform in terms of accuracy and efficiency for automated semantic modeling tasks?
    \item[\textbf{•}] \textbf{Q2:} Can the Knowledge Prompt Chaining approach effectively handle complex semantic models?
    \item[\textbf{•}] \textbf{Q3:} What is the impact of different prompt designs on the performance of semantic modeling?
\end{itemize}

\begin{table}[!htbp]
\centering
\caption{}
\label{tab:tableTab2}
\renewcommand{\arraystretch}{1.15}
\begin{tabular}{cccc}
\toprule
Datasets & Taheriyan & Serene & Ours \\
\midrule 
$ds_{crm}$ & 0.692 & 0.860 & \textbf{0.978} \\
$ds_{edm}$ & 0.797 &  0.890 & \textbf{0.991} \\
$ds_{schema}$ & 0.735 & 0.820 & \textbf{0.969} \\
\bottomrule
\end{tabular}
\end{table}

\begin{table}[ht]
\caption{}
\centering
\renewcommand{\arraystretch}{1.15}
\begin{tabular}{ccccccccc} 
\toprule 
\multirow{2}*{Datasets} & \multicolumn{4}{c}{Precision} & \multicolumn{4}{c}{Recall} \\
\cmidrule(lr){2-5} \cmidrule(lr){6-9}
 & Taheriyan & Serene & PGM-SM & Ours & Taheriyan & Serene & PGM-SM & Ours \\
\midrule
$ds_{crm}$ & 0.785 & 0.736 & 0.819 & \textbf{0.878} & 0.798 & 0.773 & 0.686 & \textbf{0.876} \\
$ds_{edm}$ & 0.858 & 0.777 & 0.752 & \textbf{0.904} & 0.856 &  0.803& 0.746 & \textbf{0.929} \\
$ds_{schema}$ & 0.683 & 0.860 & 0.671 & \textbf{0.866} & 0.618 & 0.750 & 0.602 & \textbf{0.866} \\
\bottomrule 
\end{tabular}
\end{table}

\subsection{Performance in Accuracy and Efficiency}
In this section, we evaluate the accuracy and efficiency performance of the Knowledge Prompt Chaining approach for semantic modeling.

Tables 2 and 3 present the performance of different methods on the structured data automatic semantic modeling task, where our method achieves the best scores among the three LLMs. Table 2 compares different semantic labeling methods, showing that our method outperforms others on all datasets, with semantic labeling precision reaching 96.9\% and above, far exceeding other methods. Table 3 evaluates the precision and recall performance of different automatic semantic modeling methods. On the three datasets, our method achieved accuracy improvements of 5.9\%, 4.6\%, and 0.6\%, respectively, compared to previous methods. On the $ds_{edm}$ dataset, it achieved a precision of 90.4\% and a recall of 92.9\%, indicating that the combination of LLMs and prompts is more suitable for solving complex table reasoning tasks. To further analyze the performance of the three LLMs, we conducted a detailed evaluation of the LLMs on a two-step task, with the results shown in Tables 4 and 5. The results in Table 4 show that Claude 3.5 Sonnet performs exceptionally well in the semantic annotation task. Notably, in the one-shot setting, the semantic labeling precision reaches 90.4\%, which indirectly proves that LLMs possess rich prior knowledge and demonstrate good reasoning abilities with minimal prompt input. Table 5 compares the performance of different LLMs in the semantic modeling task. It is clear that Claude 3.5 Sonnet and DeepSeek-V2.5 achieved excellent results in semantic modeling, while GPT-4 Turbo performed somewhat less effectively, indicating that different LLMs exhibit varying adaptability when facing different tasks. Table 6 presents the time required for different LLMs to predict the semantic model of a single structured data source. The results are the average values across three settings. Since network factors may affect the task prediction time when using LLMs, these data should be considered as reference. It can be observed that the prediction time for a single data point is always controlled within 30 seconds, and the smaller the data source, the shorter the prediction time.

\begin{table}[ht]
\caption{}
\centering
\renewcommand{\arraystretch}{1.15}
\begin{tabular}{cccccccc} 
\toprule 
\multirow{2}*{Datasets} & \multirow{2}*{Models} & \multicolumn{3}{c}{Precision} & \multicolumn{3}{c}{Recall} \\
\cmidrule(lr){3-5} \cmidrule(lr){6-8}
 &  & one-shot & quarter-shot & half-shot & one-shot & quarter-shot & half-shot \\
\midrule
\multirow{3}*{$ds_{crm}$} & Claude 3.5 Sonnet & \textbf{0.860} & \textbf{0.981} & \textbf{0.978} & \textbf{0.893} & \textbf{0.980} & \textbf{0.981} \\
 & GPT-4 Turbo & 0.791 & 0.966 & 0.976 & 0.798 & 0.968 & 0.978 \\
 & DeepSeek-V2.5 & 0.711 & 0.868 & 0.881 & 0.698 & 0.865 & 0.887 \\
\midrule
\multirow{3}*{$ds_{edm}$} & Claude 3.5 Sonnet & \textbf{0.847} & \textbf{0.985} & \textbf{0.991} & \textbf{0.826} & \textbf{0.982} & \textbf{0.991} \\
 & GPT-4 Turbo & 0.725 & 0.907 & 0.952 & 0.746 & 0.913 & 0.959 \\
 & DeepSeek-V2.5 & 0.730 & 0.924 & 0.931 & 0.730 & 0.924 & 0.931 \\
\midrule
\multirow{3}*{$ds_{schema}$} & Claude 3.5 Sonnet & \textbf{0.904} & \textbf{0.969} & \textbf{0.969} & \textbf{0.897} & \textbf{0.969} & \textbf{0.969} \\
 & GPT-4 Turbo & 0.760 & 0.943 & 0.948 & 0.738 & 0.943 & 0.948 \\
 & DeepSeek-V2.5 & 0.811 & 0.898 & 0.923 & 0.785 & 0.881 & 0.923 \\
\bottomrule 
\end{tabular}
\end{table}

\begin{table}[ht]
\caption{}
\centering
\renewcommand{\arraystretch}{1.15}
\begin{tabular}{cccccccc} 
\toprule 
\multirow{2}*{Datasets} & \multirow{2}*{Models} & \multicolumn{3}{c}{Precision} & \multicolumn{3}{c}{Recall} \\
\cmidrule(lr){3-5} \cmidrule(lr){6-8}
 &  & one-shot & quarter-shot & half-shot & one-shot & quarter-shot & half-shot \\
\midrule
\multirow{3}*{$ds_{crm}$} & Claude 3.5 Sonnet & 0.741 & 0.799 & 0.857 & 0.774 & 0.842 & \textbf{0.876} \\
 & GPT-4 Turbo & 0.437 & 0.752 & 0.748 & 0.392 & 0.678 & 0.722 \\
 & DeepSeek-V2.5 & \textbf{0.766} & \textbf{0.827} & \textbf{0.878} & \textbf{0.784} & \textbf{0.856} & 0.867 \\
\midrule
\multirow{3}*{$ds_{edm}$} & Claude 3.5 Sonnet & 0.578 & \textbf{0.854} & \textbf{0.904} & 0.655 & \textbf{0.885} & \textbf{0.929} \\
 & GPT-4 Turbo & 0.581 & 0.711 & 0.819 & 0.587 & 0.711 & 0.784 \\
 & DeepSeek-V2.5 & \textbf{0.645} & 0.847 & 0.879 & \textbf{0.716} & 0.866 & 0.891 \\
\midrule
\multirow{3}*{$ds_{schema}$} & Claude 3.5 Sonnet & \textbf{0.838} & \textbf{0.907} & \textbf{0.866} & \textbf{0.698} & \textbf{0.907} & \textbf{0.866} \\
 & GPT-4 Turbo & 0.760 & 0.853 & 0.855 & 0.575 & 0.825 & 0.827 \\
 & DeepSeek-V2.5 & 0.500 & 0.667 & 0.750 & 0.667 & 0.667 & 0.722 \\
\bottomrule 
\end{tabular}
\end{table}

\begin{table}[ht]
\caption{}
\centering
\renewcommand{\arraystretch}{1.15}
\begin{tabular}{cccc} 
\toprule 
Datasets & Claude 3.5 Sonnet & GPT-4 Turbo & DeepSeek-V2.5 \\
\midrule
$ds_{crm}$ & 21s & 26s & 20s \\
$ds_{edm}$ & 15s & 18s & 12s \\
$ds_{schema}$ & 8s & 11s & 7s \\
\bottomrule 
\end{tabular}
\end{table}

\begin{table}[ht]
\caption{}
\centering
\renewcommand{\arraystretch}{1.15}
\begin{tabular}{ccccc} 
\toprule 
\multirow{2}*{Datasets} & \multicolumn{2}{c}{$ds_{crm}$} & \multicolumn{2}{c}{$ds_{edm}$} \\
\cmidrule(lr){2-3} \cmidrule(lr){4-5}
 & Precision & Recall & Precision& Recall  \\
\midrule
Ablation + Prompting Chaining + Prune & 0.878 &	0.867 &	0.862	&0.911 \\
Ablation + Prompting Chaining & 0.860	&0.860 &	0.858	&0.911 \\
Ablation & 0.618 &0.639	&0.683&	0.761 \\
\bottomrule 
\end{tabular}
\end{table}

\begin{figure}[t]
\includegraphics[scale=0.19]{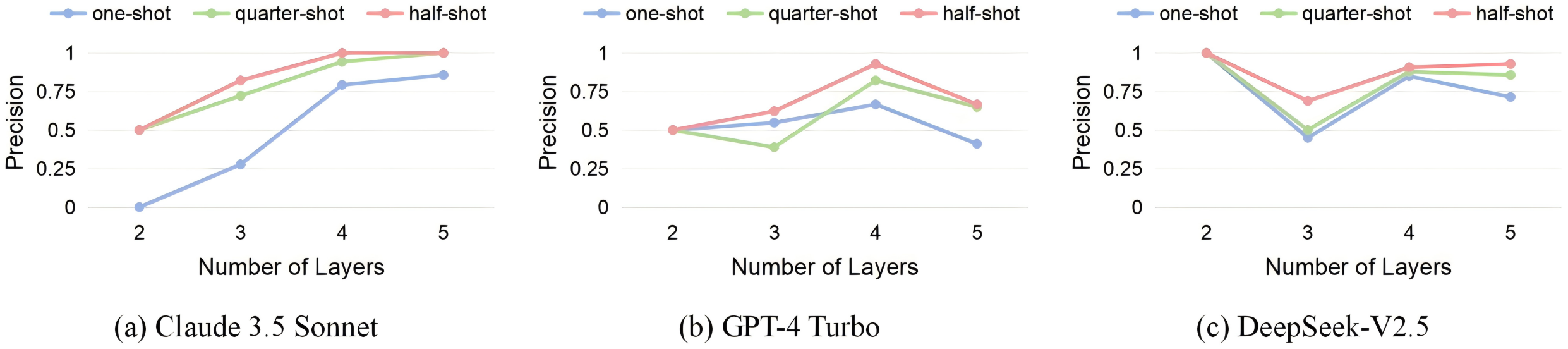}
\centering
\caption{Linear Relationship Between Complexity of Semantics and Performance Score}
\label{Complexity}
\end{figure}

\subsection{Performance on Complex Semantic Models}
We evaluate Knowledge Prompt Chaining framework's performance on different levels of graph complexity. We use the depth of graphs in the semantic models to measure the complexity of the graph. Claude's model shows high performance on complicated graphs while not doing well on the simplest graphs. DeepSeek has a good performance in all levels of complexity especially in half-shot set. GPT-4 Turbo shows a pretty good performance in 4-depth graphs and also has the lowest precision score in the simplest graphs. The results in Figure~\ref{Complexity} shows our method's strong ability to deal with complicated graphs. It implied that Claude and GPT-4 might have some issues with "overthinking" since they can not deal with the simplest graphs well.

\subsection{Effectiveness Analysis of Prompt Ablation Designs}
We analyze different parts of our framework by evaluating their contributions to the performance. We use the DeepSeek model to test them on CRM and EDM datasets. We start from an ablation framework and add components one by one. As shown in Table 7, our method surpasses the ablation version across all data. The prompting chaining part in our framework contributes to the most prat of performance improvement. And pruning technique also works well and helps the framework reduce hallucinations of the LLM.

\section{Conclusions and Future Work}
In this paper, we introduce the knowledge prompting chaining framework to automatically generate semantic models for structured data. By implementing knowledge injection and prompting chaining and graph pruning, our method improves the accuracy of the task across different datasets without any training requirement.

In the future, we will explore the prompting turning methods to further enhance the reasoning ability of our framework. The reasoning details in $<Step1>$ and $<Step2>$ that we require the model to generate can be treated as synthetic data. Additionally, in this paper, we designed $<Rule>$ part manually. If we can extract this part automatically, the efficiency of our framework should be enhanced. In this aspect, we aim to apply meta prompt\cite{zhang2023meta} to let LLM teach themselves to find $<Rule>$.
\bibliographystyle{unsrt}
\bibliography{paper}

\begin{thebibliography}{10}

\bibitem{dhamankar2004imap}
Robin Dhamankar, Yoonkyong Lee, AnHai Doan, Alon Halevy, and Pedro Domingos.
\newblock imap: Discovering complex semantic matches between database schemas.
\newblock In {\em Proceedings of the 2004 ACM SIGMOD international conference on Management of data}, pages 383--394, 2004.

\bibitem{Taheriyan2016LearningTS}
Mohsen Taheriyan, Craig~A. Knoblock, Pedro~A. Szekely, and J.~Ambite.
\newblock Learning the semantics of structured data sources.
\newblock {\em J. Web Semant.}, 37-38:152--169, 2016.

\bibitem{de2018machine}
Diego De~U{\~n}a, Nataliia R{\"u}mmele, Graeme Gange, Peter Schachte, and Peter~J Stuckey.
\newblock Machine learning and constraint programming for relational-to-ontology schema mapping.
\newblock In {\em International Joint Conference on Artificial Intelligence 2018}, pages 1277--1283. Association for the Advancement of Artificial Intelligence (AAAI), 2018.

\bibitem{10.1145/3308558.3313711}
Binh Vu, Craig Knoblock, and Jay Pujara.
\newblock Learning semantic models of data sources using probabilistic graphical models.
\newblock In {\em The World Wide Web Conference}, WWW '19, page 1944–1953, New York, NY, USA, 2019. Association for Computing Machinery.

\bibitem{futia2020semi}
Giuseppe Futia, Antonio Vetr{\`o}, and Juan~Carlos De~Martin.
\newblock Semi: A semantic modeling machine to build knowledge graphs with graph neural networks.
\newblock {\em SoftwareX}, 12:100516, 2020.

\bibitem{pham2016semantic}
Minh Pham, Suresh Alse, Craig~A Knoblock, and Pedro Szekely.
\newblock Semantic labeling: a domain-independent approach.
\newblock In {\em The Semantic Web--ISWC 2016: 15th International Semantic Web Conference, Kobe, Japan, October 17--21, 2016, Proceedings, Part I 15}, pages 446--462. Springer, 2016.

\bibitem{rummele2018evaluating}
Natalia R{\"u}mmele, Yuriy Tyshetskiy, and Alex Collins.
\newblock Evaluating approaches for supervised semantic labeling.
\newblock {\em arXiv preprint arXiv:1801.09788}, 2018.

\bibitem{takeoka2019meimei}
Kunihiro Takeoka, M.~Oyamada, Shinji Nakadai, and Takeshi Okadome.
\newblock Meimei: An efficient probabilistic approach for semantically annotating tables.
\newblock In {\em AAAI Conference on Artificial Intelligence}, 2019.

\bibitem{jaitly2023towards}
Sukriti Jaitly, Tanay Shah, Ashish Shugani, and Razik~Singh Grewal.
\newblock Towards better serialization of tabular data for few-shot classification.
\newblock {\em arXiv preprint arXiv:2312.12464}, 2023.

\bibitem{singha2023tabular}
Ananya Singha, José Cambronero, Sumit Gulwani, Vu~Le, and Chris Parnin.
\newblock Tabular representation, noisy operators, and impacts on table structure understanding tasks in llms.
\newblock In {\em Table Representation Learning Workshop at NeurIPS 2023}, December 2023.

\bibitem{hegselmann2023tabllm}
Stefan Hegselmann, Alejandro Buendia, Hunter Lang, Monica Agrawal, Xiaoyi Jiang, and David Sontag.
\newblock Tabllm: Few-shot classification of tabular data with large language models.
\newblock In {\em International Conference on Artificial Intelligence and Statistics}, pages 5549--5581. PMLR, 2023.

\bibitem{sui2023tap4llm}
Yuan Sui, Jiaru Zou, Mengyu Zhou, Xinyi He, Lun Du, Shi Han, and Dongmei Zhang.
\newblock Tap4llm: Table provider on sampling, augmenting, and packing semi-structured data for large language model reasoning.
\newblock {\em arXiv preprint arXiv:2312.09039}, 2023.

\bibitem{borisov2022language}
Vadim Borisov, Kathrin Se{\ss}ler, Tobias Leemann, Martin Pawelczyk, and Gjergji Kasneci.
\newblock Language models are realistic tabular data generators.
\newblock {\em arXiv preprint arXiv:2210.06280}, 2022.

\bibitem{wei2022chain}
Jason Wei, Xuezhi Wang, Dale Schuurmans, Maarten Bosma, Fei Xia, Ed~Chi, Quoc~V Le, Denny Zhou, et~al.
\newblock Chain-of-thought prompting elicits reasoning in large language models.
\newblock {\em Advances in neural information processing systems}, 35:24824--24837, 2022.

\bibitem{zhao2023large}
Bowen Zhao, Changkai Ji, Yuejie Zhang, Wen He, Yingwen Wang, Qing Wang, Rui Feng, and Xiaobo Zhang.
\newblock Large language models are complex table parsers.
\newblock {\em arXiv preprint arXiv:2312.11521}, 2023.

\bibitem{wang2024chain}
Zilong Wang, Hao Zhang, Chun-Liang Li, Julian~Martin Eisenschlos, Vincent Perot, Zifeng Wang, Lesly Miculicich, Yasuhisa Fujii, Jingbo Shang, Chen-Yu Lee, et~al.
\newblock Chain-of-table: Evolving tables in the reasoning chain for table understanding.
\newblock {\em arXiv preprint arXiv:2401.04398}, 2024.

\bibitem{wei2024kicgpt}
Yanbin Wei, Qiushi Huang, James~T Kwok, and Yu~Zhang.
\newblock Kicgpt: Large language model with knowledge in context for knowledge graph completion.
\newblock {\em arXiv preprint arXiv:2402.02389}, 2024.

\bibitem{shu2024knowledge}
Dong Shu, Tianle Chen, Mingyu Jin, Chong Zhang, Mengnan Du, and Yongfeng Zhang.
\newblock Knowledge graph large language model (kg-llm) for link prediction.
\newblock {\em arXiv preprint arXiv:2403.07311}, 2024.

\bibitem{wang2024soft}
Qunbo Wang, Ruyi Ji, Tianhao Peng, Wenjun Wu, Zechao Li, and Jing Liu.
\newblock Soft knowledge prompt: Help external knowledge become a better teacher to instruct llm in knowledge-based vqa.
\newblock In {\em Proceedings of the 62nd Annual Meeting of the Association for Computational Linguistics (Volume 1: Long Papers)}, pages 6132--6143, 2024.

\bibitem{fang2024large}
Xi~Fang, Weijie Xu, Fiona~Anting Tan, Ziqing Hu, Jiani Zhang, Yanjun Qi, Srinivasan~H. Sengamedu, and Christos Faloutsos.
\newblock Large language models ({LLM}s) on tabular data: Prediction, generation, and understanding - a survey.
\newblock {\em Transactions on Machine Learning Research}, 2024.

\bibitem{zhang2023meta}
Yifan Zhang, Yang Yuan, and Andrew Chi-Chih Yao.
\newblock Meta prompting for agi systems.
\newblock {\em arXiv preprint arXiv:2311.11482}, 2023.

\end{thebibliography}

\begin{appendix}
\renewcommand{\thesection}{\Alph{section}}
\section*{APPENDIX}
\section{Updates of Semantic Models}
The semantic model updates in this study are applied exclusively to the $ds_{crm}$ dataset. Previous work revealed certain limitations in the provided gold standard model, such as meaningless leaf nodes, a lack of interpretability in the alignment between the semantic model and the data source, and inconsistent standards in semantic model construction. Guided by domain experts, we updated the gold standard model to ensure it is more logical and precise. The updates adhere to the following principles:
\begin{enumerate}[leftmargin=*, labelwidth=0pt, labelsep=0.5em]
    \item Removal of Irrelevant Leaf Nodes: Based on the table headers provided in the 28 data sources, we removed leaf nodes from the model that were not directly associated with any table headers. All leaf nodes in the updated model must correspond to a table header in the data source, with names strictly aligned to ensure consistency. This approach ensures that every leaf node in the model is meaningful and aligned with the data source.
    \item Numbering for Multiple \texttt{crm:E52\_Time-Span} Nodes: When multiple \texttt{crm:E52\_Time-Span} nodes exist in the model, the following numbering rules are applied.
    \begin{itemize}
        \item If \texttt{crm:E67\_Birth} and \texttt{crm:E69\_Death} nodes are present, the nodes are numbered as <\texttt{crm:E67\_Birth1}, \texttt{crm:P4\_has\_time-span}, \texttt{crm:E52\_Time-Span1}> and <\texttt{crm:E69\_Death1}, \texttt{crm:P4\_has\_time-span}, \texttt{crm:E52\_Time-Span2}>. Any remaining \texttt{crm:E52\_Time-Span} nodes are numbered sequentially afterward.
        \item If \texttt{crm:E67\_Birth} and \texttt{crm:E69\_Death} nodes are absent, all \texttt{crm:E52\_Time-Span} nodes are numbered without specific constraints.
    \end{itemize}
    \item Usage Guidelines for \texttt{crm:E21\_Person} and \texttt{crm:E39\_Actor} Nodes: The choice between \texttt{crm:E21\_Person} and \texttt{crm:E39\_Actor} nodes depends on the context of the data source.
    \begin{itemize}
        \item If the data source primarily focuses on personal information about artists without referencing their works, the \texttt{crm:E39\_Actor} node is used.
        \item In all other cases, the \texttt{crm:E21\_Person} node is applied.
    \end{itemize}
    \item Guidelines for the Use of \texttt{crm:E55\_Type} Nodes: Two categories of \texttt{crm:E55\_Type} nodes are employed across all models, differentiated based on the types referenced in the data source records.
    \begin{itemize}
        \item The first category connects to \texttt{crm:E22\_Man-Made\_Object} via <\texttt{crm:E22\_Man-Made\_Object}, \texttt{crm:P2\_has\_type}, \texttt{crm:E55\_Type}>. These \texttt{crm:E55\_Type} nodes are further linked to table headers representing broad artistic types. The records in these table headers may include values such as “Painting”, “Furniture”, or “Architecture”, which represent broad artistic categories.
        \item The second category connects to \texttt{crm:E12\_Production} via <\texttt{crm:E12\_Production}, \texttt{crm:P32\_used\_general\_technique}, \texttt{crm:E55\_Type}>. These \texttt{crm:E55\_Type} nodes are further linked to table headers representing detailed production techniques or material types. The records in these table headers may include values such as “Oil on canvas”, “Mahogany, chestnut, tulip poplar”, or “Wood”, which represent detailed categories.
    \end{itemize}
\end{enumerate}
\end{appendix}







\end{document}